\documentclass{article}
\usepackage{graphicx}

\begin{document}

\title{V-like Formations in Flocks of Artificial Birds}

\author{Andre~Nathan\\
Valmir~C.~Barbosa\thanks{Corresponding author (valmir@cos.ufrj.br).}\\
\\
Universidade Federal do Rio de Janeiro\\
Programa de Engenharia de Sistemas e Computa\c c\~ao, COPPE\\
Caixa Postal 68511\\
21941-972 Rio de Janeiro - RJ, Brazil}

\date{}

\maketitle

\begin{abstract}
We consider flocks of artificial birds and study the emergence of V-like
formations during flight. We introduce a small set of fully distributed
positioning rules to guide the birds' movements and demonstrate, by means of
simulations, that they tend to lead to stabilization into several of the
well-known V-like formations that have been observed in nature. We also provide
quantitative indicators that we believe are closely related to achieving V-like
formations, and study their behavior over a large set of independent
simulations.

\end{abstract}

\newpage
\section{Introduction}

Migratory birds in flight often group into V-like formations, that is, they tend
to self-organize into a flight formation having very few individuals in lead
positions while the others group behind them near the edges of imaginary
trailing V's whose vertices are the lead birds. While it seems to be a consensus
that group flight provides greater protection against predators, the reasons for
the particular grouping into laterally slanted, nearly straight lines have
elicited two distinct lines of explanation, following the seminal works in
\cite{lissaman70} and \cite{gould74}, respectively.

The first one has evoked the aerodynamics of bird flight, whose essence is that
each flying individual creates an upwash region behind it, just off the tips
of its wings, such that another individual benefits greatly (in terms of
requiring less exertion during flight) if it places one of its wings in that
region. For relatively large birds, which are the ones that have been observed
to group into V-like formations most often, this seems to be the main reason for
formation flight
\cite{hummel83, hainsworth87, badgerow88, speakman98, weimerskirch01}. The
second, competing line of reasoning has been that flying in a somewhat skewed
position relative to the bird in front is crucial for an individual's
orientation, in addition to allowing unhindered visual communication and
therefore helping avoid collisions. For relatively small birds, the aerodynamic
benefits are less relevant and, moreover, the observed lateral and longitudinal
separations between nearest birds seem to be correlated in a way that strongly
supports this vision-related explanation \cite{cutts94}.

In this study, we consider simple artificial birds and address the question of
whether positioning rules exist that, during flight, can lead a flock to settle
into some stable V-like formation. Such rules should be founded on a blend of
the basic explanatory trends of placing more weight on aerodynamics- or
vision-related justifications, thus consonant with the current belief that the
traits that allow natural birds to take advantage of one or the other benefit
evolved concomitantly \cite{rayner01}. They should, in addition, be inherently
distributed, so that only sensory information is to guide each bird's course of
action, and also motivated by the birds' innate drive at flying as a flock. As
we demonstrate in the remainder of this letter, rules with these characteristics
do exist that are both robust (in the sense of ensuring that stable
configurations can be expected to be reached) and flexible (allowing for a rich
variety of V-like formations).

Our work is preceded by important related contributions, notably by the simple
rules in \cite{reynolds87}, which in essence say that birds should avoid
colliding with one another while attempting to maintain the same pace as the
nearest birds as well as seeking to be positioned as near their center as
possible. Such rules are not meant to give rise to V-like formations, only to
the flight as a group itself. Adding to them the further rule that a bird should
strive to keep some portion of the visual field unobstructed is reported to have
caused some of the birds to coalesce into fragmented V-like formations
\cite{flake98}, but the whole approach seems to have remained oblivious to the
sensory input that birds derive from favorable aerodynamics. We also mention the
fuzzy rules introduced in \cite{bajec03b,bajec05}, but they fail to induce
V-like formations, except in the case of very special initial conditions. Less
related to this study but very relevant nonetheless are mathematical analyses of
flocking, both in the sense of \cite{reynolds87} (cf., for example,
\cite{toner98} and \cite{olfati-saber06}, the latter partly based on
\cite{olfati-saber04}) and under the assumption of a V-formation (e.g.,
\cite{sugimoto03}).

\section{Positioning rules}

Our own set of rules is very succinct, comprising only three rules, each
related to one of the three guiding principles discussed earlier (birds should
flock, be afforded some unobstructed view in the direction of flight, and
benefit from regions of upwash). They are the following.
\begin{description}
\item[Rule 1 (coalescing rule):] Seek the proximity of the nearest bird.
\item[Rule 2 (gap-seeking rule):] If Rule 1 is no longer applicable, seek the
nearest position that affords an unobstructed longitudinal view.
\item[Rule 3 (stationing rule):] Apply Rule 2 while the view that is sought is
not obtained or the effort to keep up with the group decreases due to increased
upwash.
\end{description}

These three rules are an informal expression of our understanding of how each of
the guiding principles is to influence a bird's actions. First of all, Rules~1
and~2 imply that we make a distinction between two modes of behavior, one
succeeding, or sometimes alternating with, the other: birds are to seek first
the closeness of the group, and only then being afforded some clear view.
Secondly, Rules~2 and~3 imply that the sensory input that keeps the bird moving
toward a relative position of clear view is not only visual but also originates
in the ease with which it is keeping up with the group.

Rules~1--3 require further specification before we describe our computational
experiments and their results. Note, first, that both Rule~1 and Rule~2 make
implicit reference to the region of space where some entity is to be sought
(the nearest bird in Rule~1, the nearest position with a clear view in Rule~2).
We assume, in both cases, that this region is delimited, on the plane of
movement and both to the right and left of the direction of movement, by an
angle $\alpha/2$. We assume also that $\alpha\le 180^\circ$ (this is in spite
of the fact that many birds in nature have visual fields delimited by larger
angles, since $\alpha$ does not delimit a bird's visual field but the region
where it is to seek the input needed by Rules~1 and~2).

\begin{figure}[t]
\centering
\scalebox{1.000}{\includegraphics{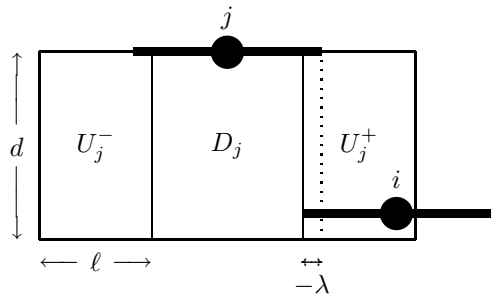}}
\caption{A flying bird $j$'s upwash ($U_j^-$ and $U_j^+$) and downwash ($D_j$)
regions. Bird $i$ is positioned for maximum upwash.}
\label{fig:wash}
\end{figure}

In a similar vein, Rules~1 and~3 allude, also implicitly, to the existence of a
closed region surrounding each bird inside which proximity is attained (Rule~1)
or upwash is found (Rule~3). Given the aforementioned constraints on $\alpha$,
and disregarding as negligible the upwash that some authors claim may be
generated by a flying bird even in the space just ahead of it
\cite{sugimoto03,anderson04}, we assume that this region is as follows, with
reference to Figure~\ref{fig:wash} [in this figure, as in others to come, an
artificial bird is depicted as a filled circle (the bird's body) with two
protruding straight-line segments (its wings)]. If $i$ is the bird seeking the
proximity of another (say $j$) in Rule~1, then proximity is attained if the body
of $i$ is in $U_j^-\cup D_j\cup U_j^+$. As for finding upwash in Rule~3, all
that is needed is that any portion of bird $i$, body or a wing segment of any
size, be in $U_j^-\cup U_j^+$ while no portion is in $D_j$. In this case, the
optimal relative placement of $i$ and $j$ (i.e., the relative placement for
maximum upwash) occurs when the lateral separation between them is
$\lambda=(\pi/4-1)w/2\approx -0.1073w$, where $w$ is a bird's wingspan, assumed
the same for all birds \cite{hummel83,hummel95,seiler02} (the birds'
longitudinal separation inside $U_j^-\cup U_j^+$ is thought to be irrelevant for
maximum upwash \cite{lissaman70,hummel83,seiler02}). So, for optimality, $i$ and
$j$ must have laterally overlapping wings.

\section{An algorithm and computational results}

Our simulation algorithm operates on $n$ birds and runs from some initial
positioning of the birds inside a unit square in two-dimensional space. Birds
are in no way constrained to flying inside this unit square, even though by
Rule~1 it seems reasonable to expect that a fixed-size square exists which, at
all times, encompasses all birds. They may then be thought of as inhabiting
unbounded two-dimensional space.

We assume that the lateral and longitudinal directions are parallel to the
original unit square's sides. In order to move according to Rules~1--3, a bird's
position relative to the others' must undergo lateral as well as longitudinal
displacements. Except for the occurrence of these displacements, we assume that
birds have no lateral speed and that they all fly at the same longitudinal
speed, giving rise to a common velocity vector that we henceforth refer to as
the baseline velocity. Of course, each bird's actual velocity vector at a given
instant is the sum of the baseline velocity and the vector comprising the bird's
lateral and longitudinal rates of displacement at that instant.

\begin{figure}[p]
\centering
\begin{tabular}{c@{\hspace{0.20in}}c}
\scalebox{0.500}{\includegraphics{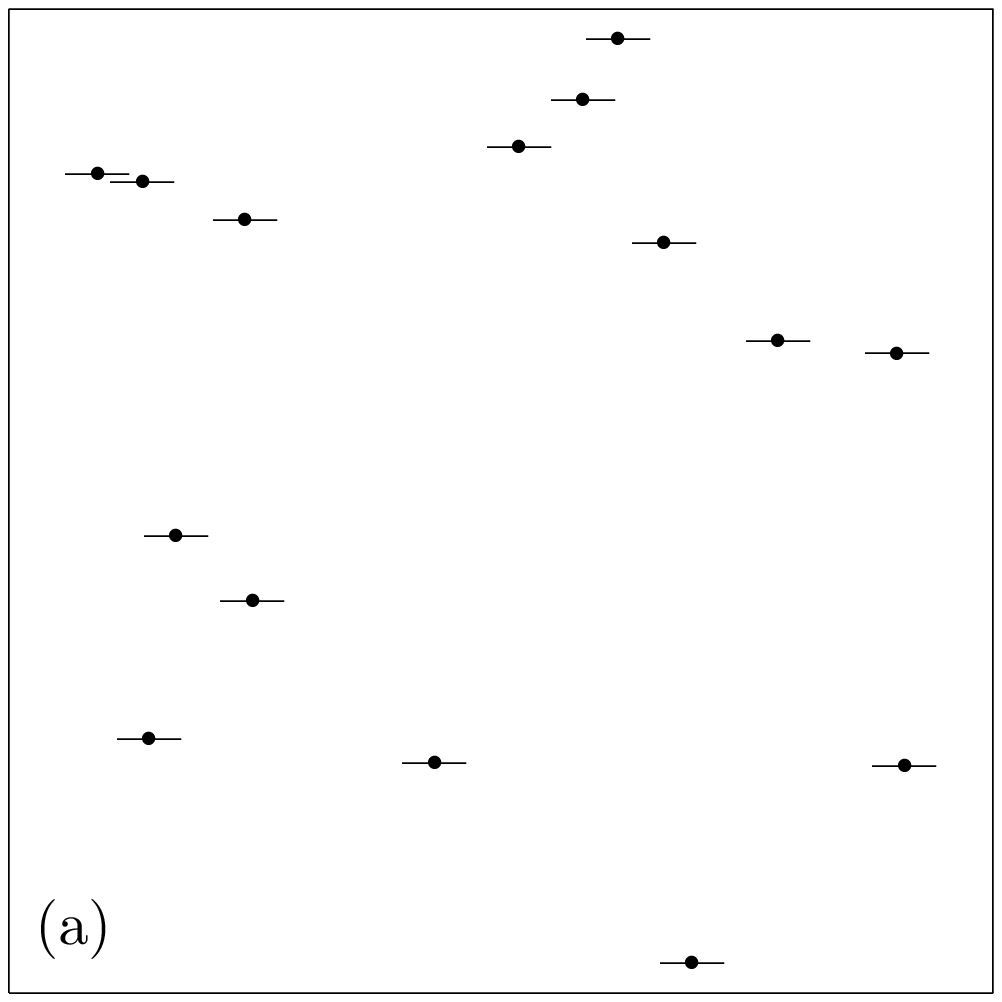}}&
\scalebox{0.500}{\includegraphics{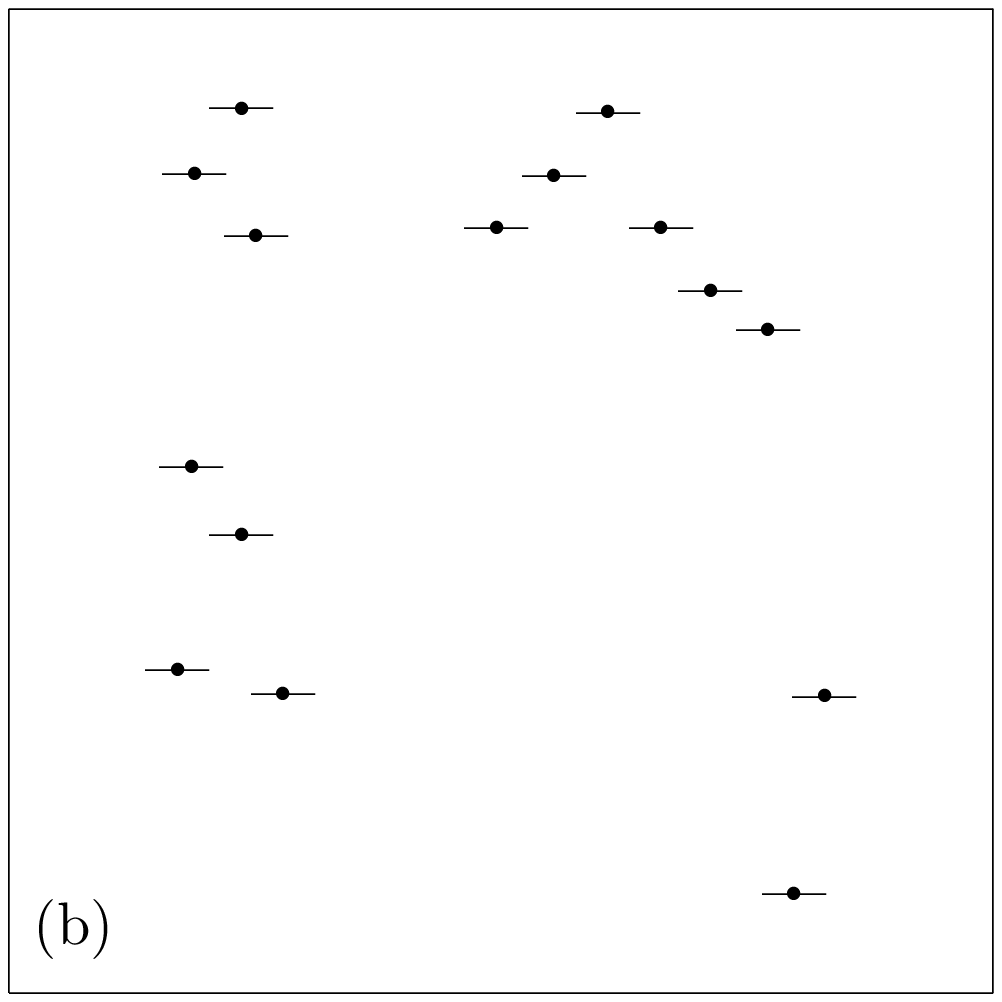}}\\
\scalebox{0.500}{\includegraphics{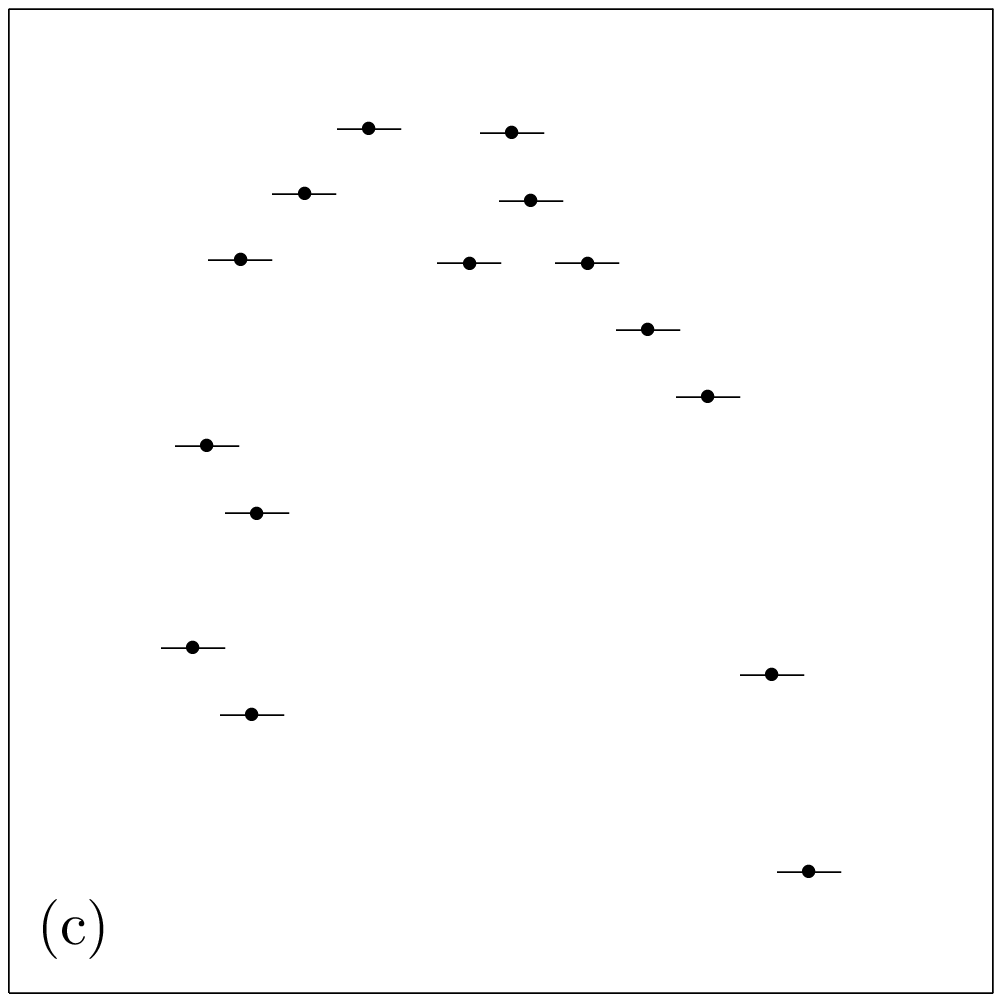}}&
\scalebox{0.500}{\includegraphics{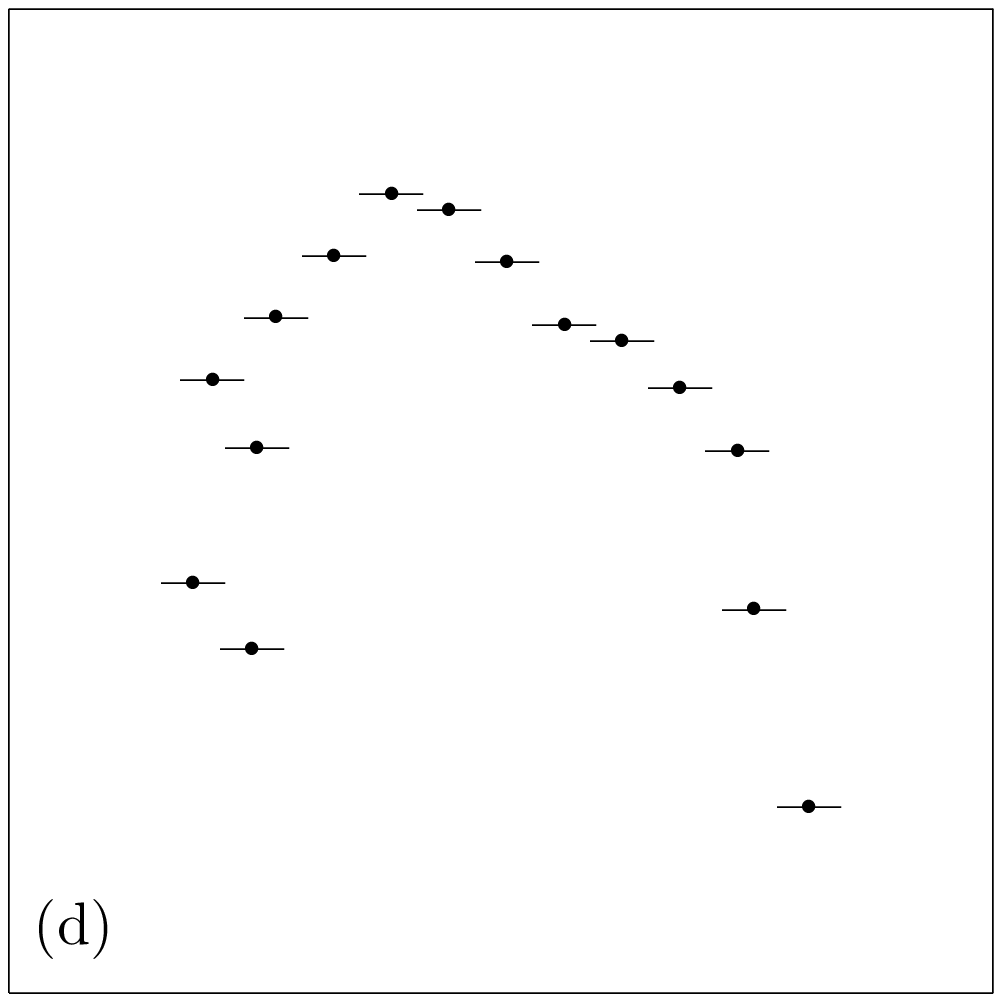}}\\
\scalebox{0.500}{\includegraphics{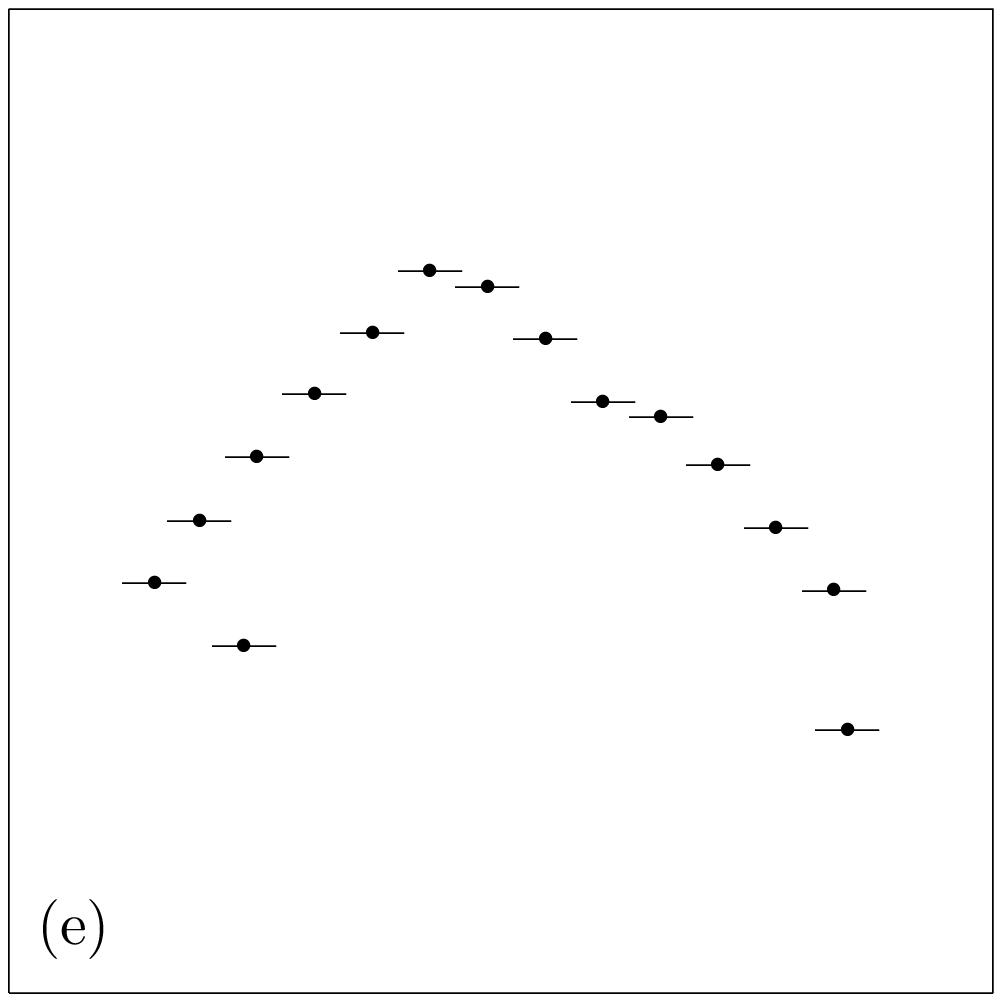}}&
\scalebox{0.500}{\includegraphics{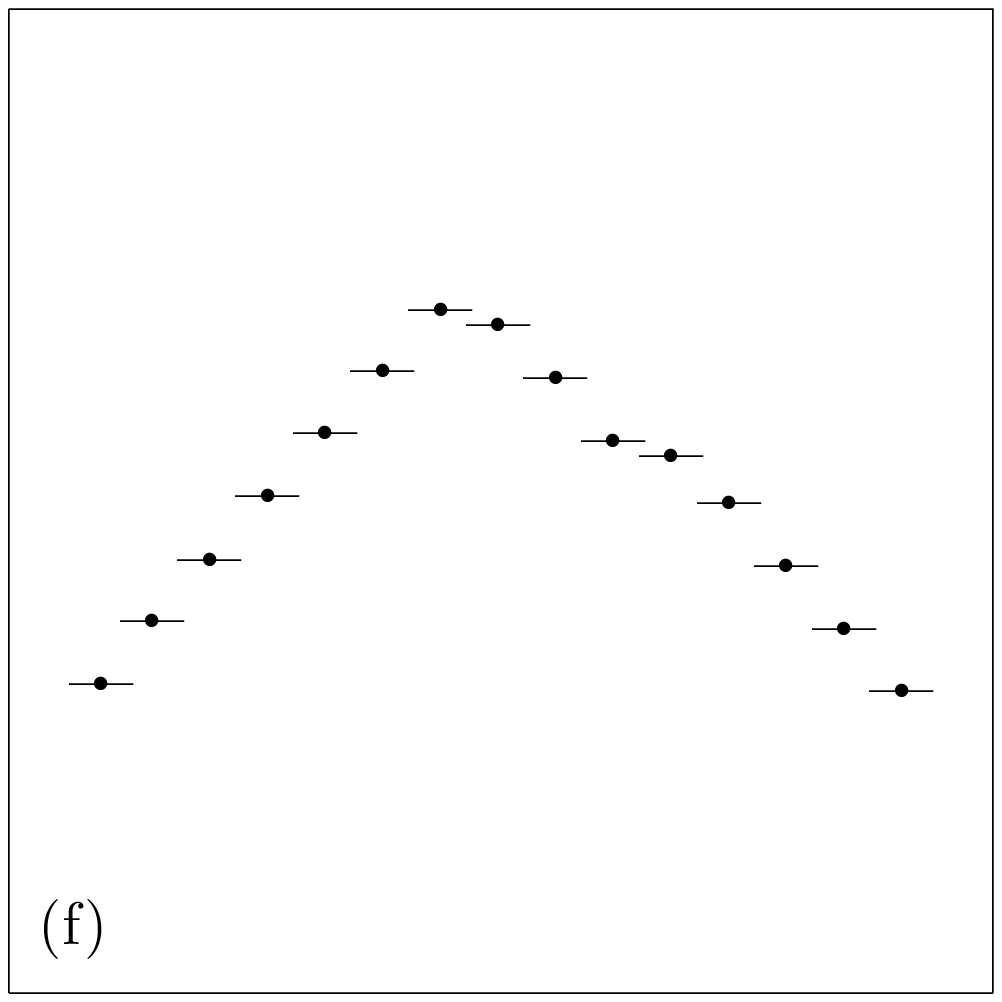}}\\
\end{tabular}
\caption{Sample evolution of relative positions for $n=15$. Time increases from
(a) to (b) by $40$ time steps, from (b) to (c) by another $40$, and so on.}
\label{fig:snap}
\end{figure}

\begin{figure}[p]
\centering
\begin{tabular}{c@{\hspace{0.20in}}c}
\scalebox{0.500}{\includegraphics{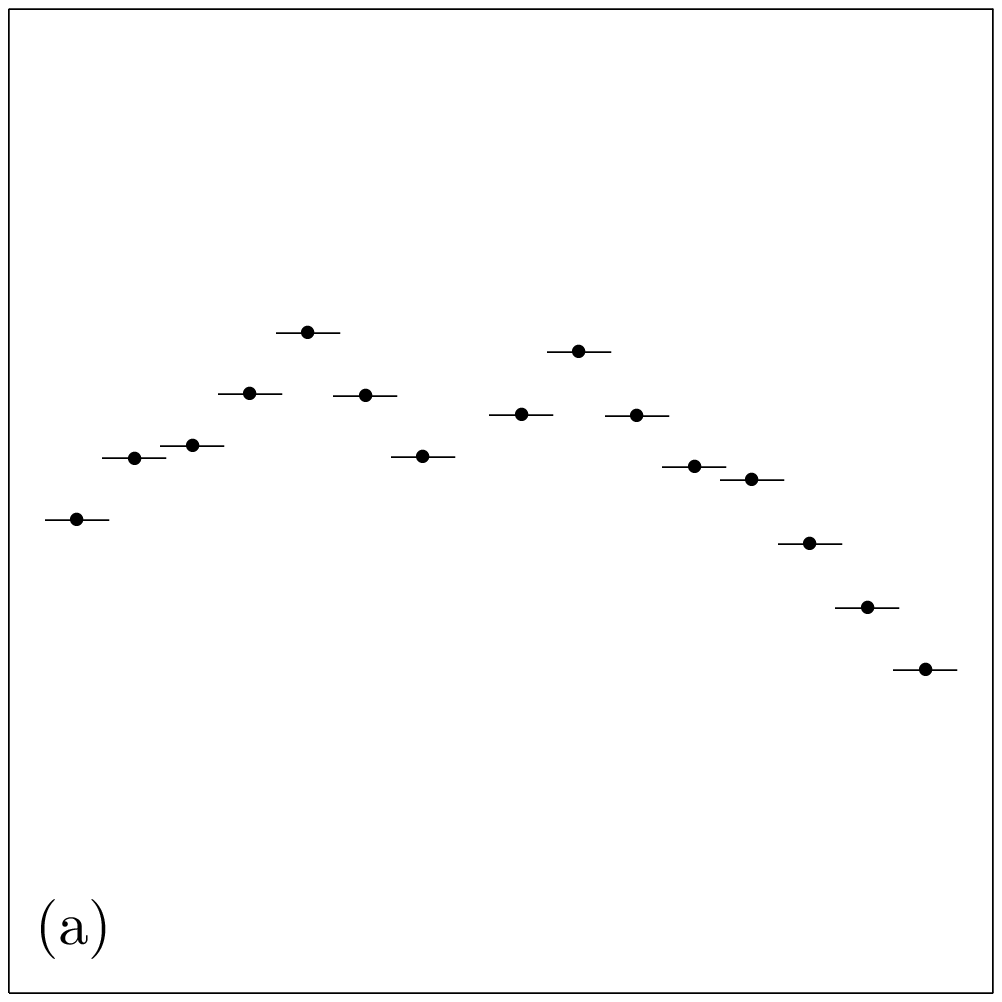}}&
\scalebox{0.500}{\includegraphics{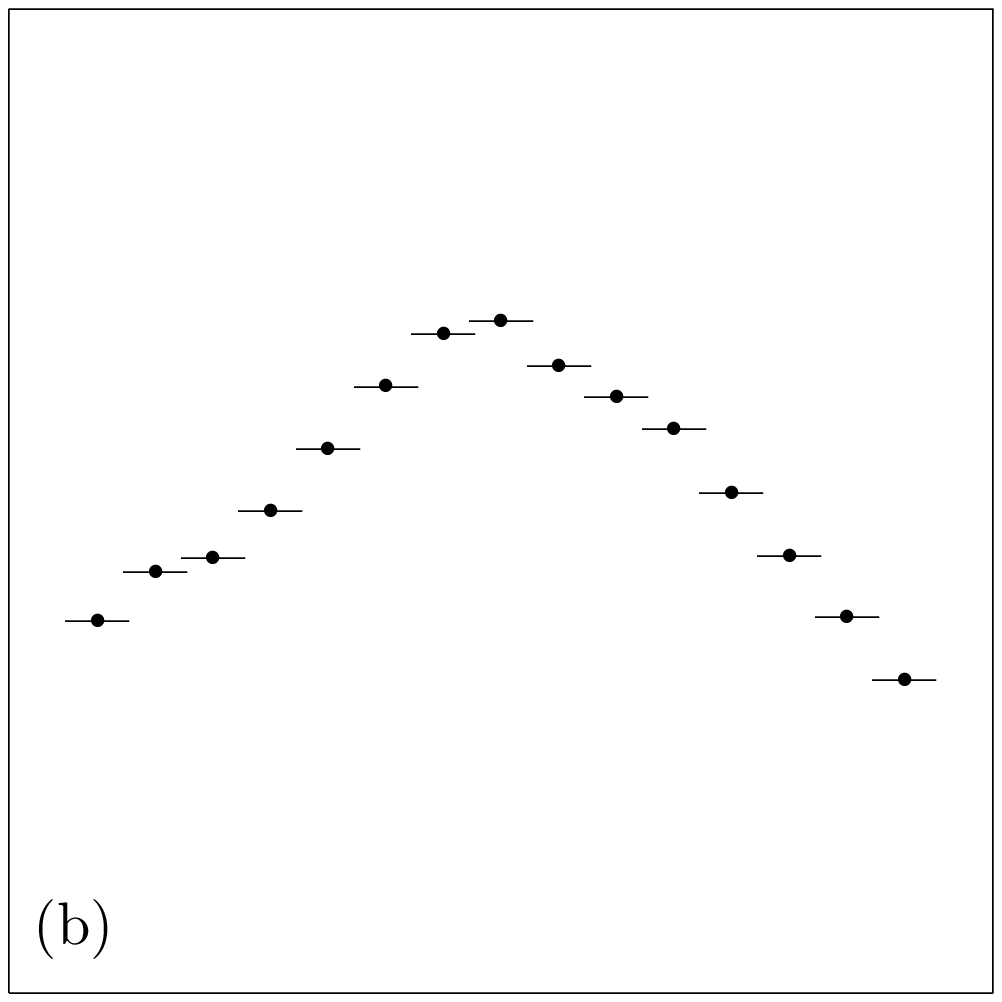}}\\
\scalebox{0.500}{\includegraphics{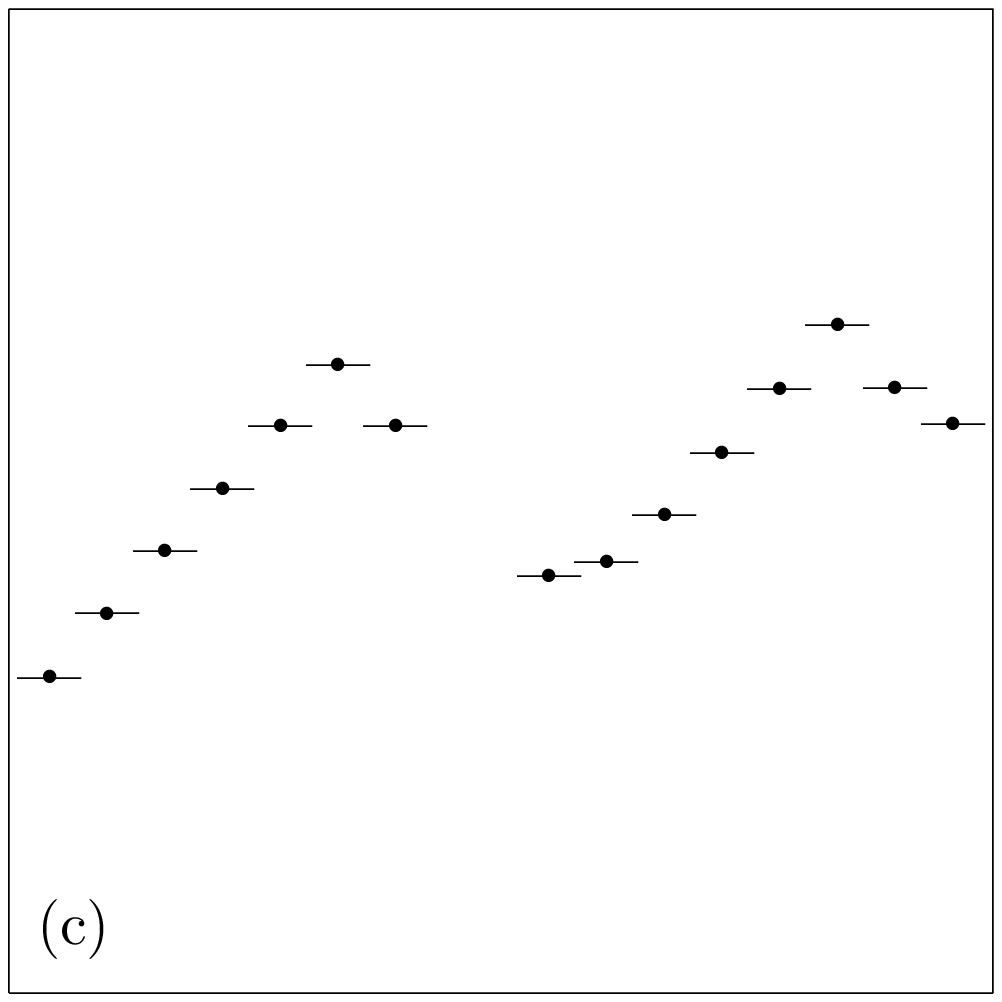}}&
\scalebox{0.500}{\includegraphics{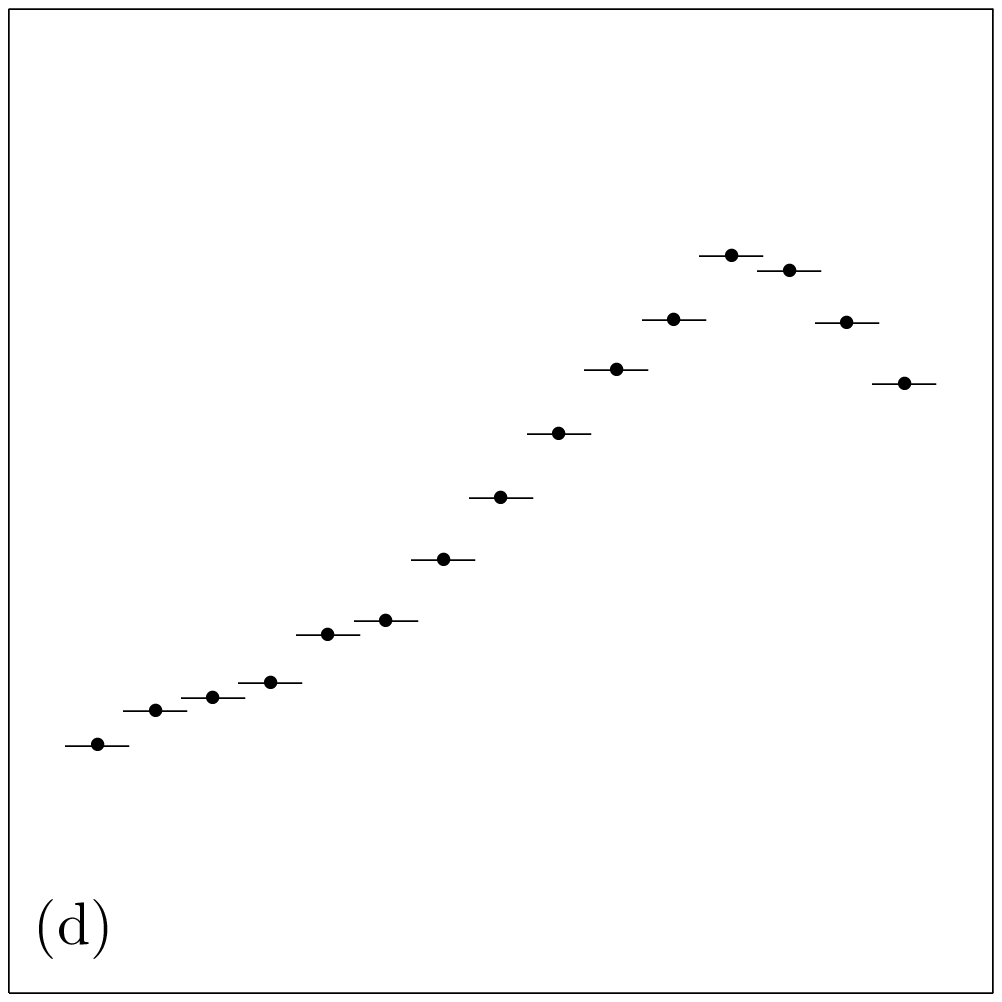}}\\
\scalebox{0.500}{\includegraphics{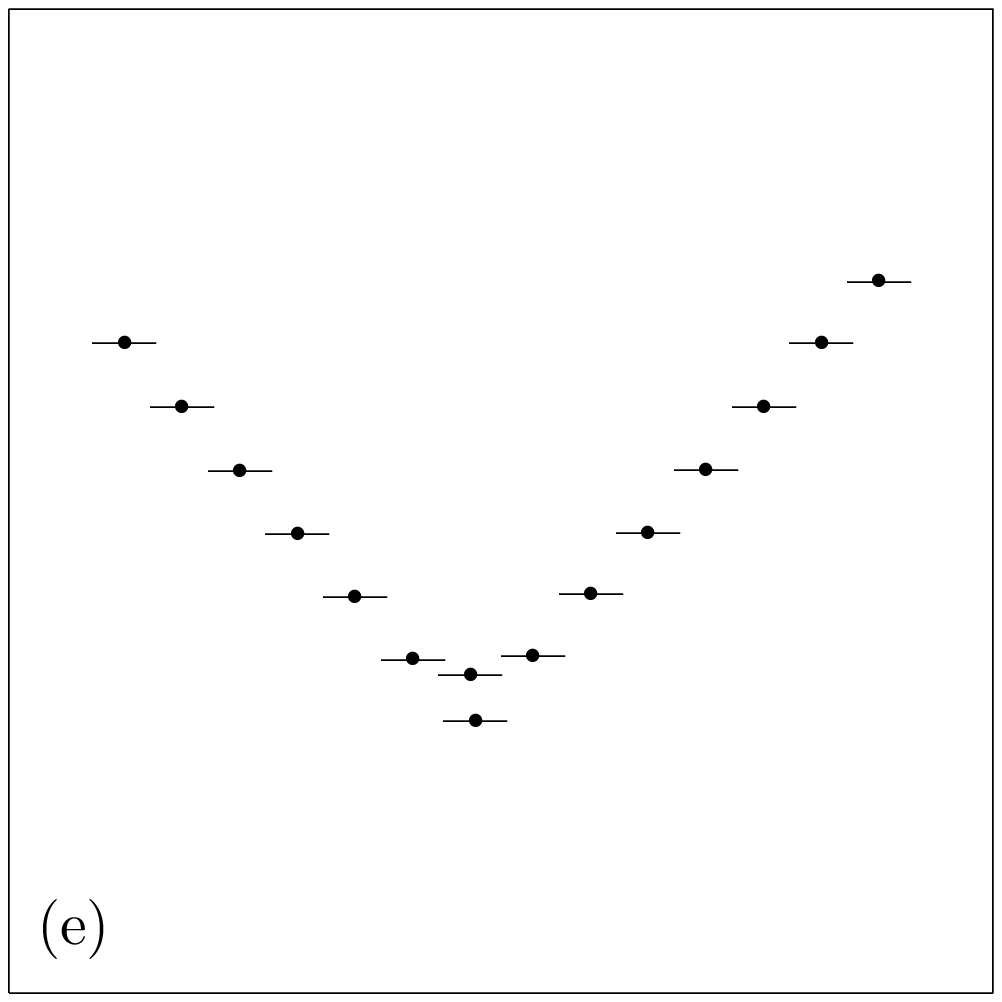}}&
\scalebox{0.500}{\includegraphics{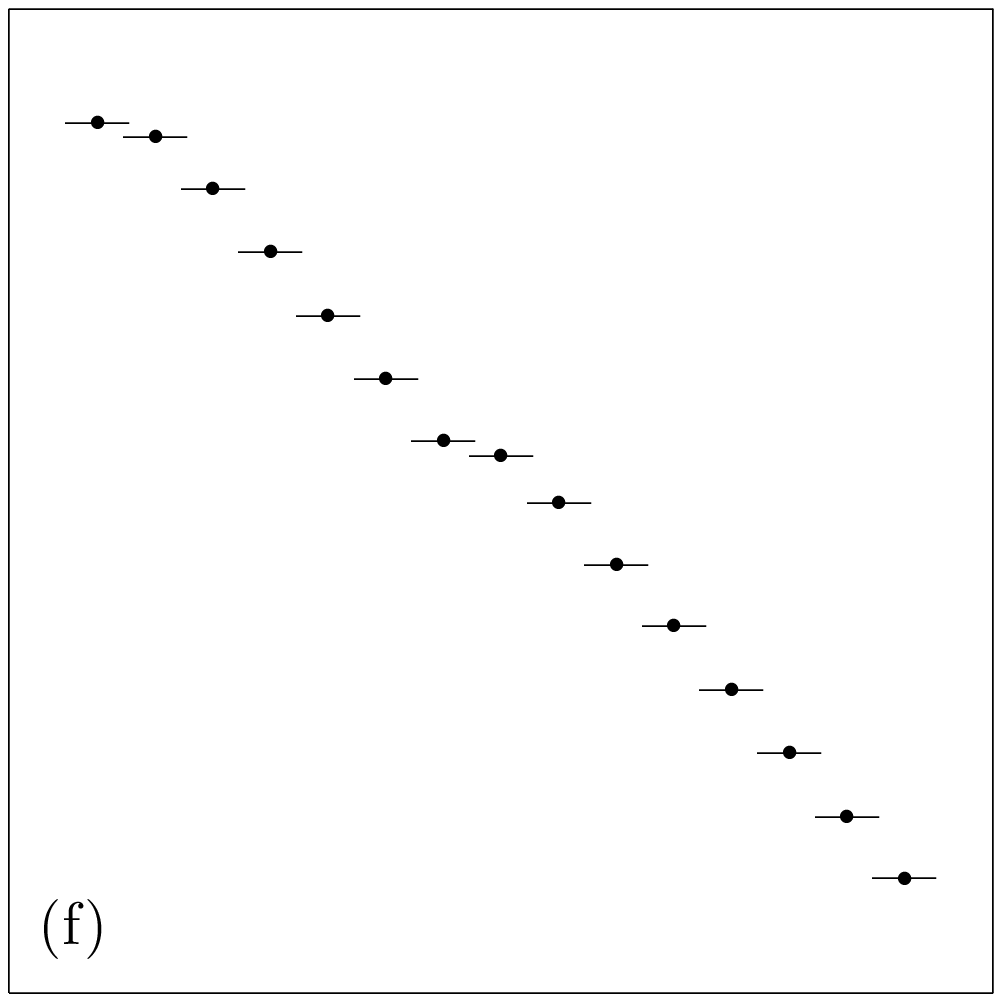}}\\
\end{tabular}
\caption{Sample formations after $T=2\,000$ time steps from independent, random
initial placements for $n=15$, with $\alpha=170^\circ$ (a, c, e) and
$\alpha=180^\circ$ (b, d, f). Samples include the celebrated W (a) and V (b)
formations, in addition to a formation into multiple unconnected groups (c) and
those formations that in \cite{heppner74} have been called the J (d), the
inverted V (e), and the echelon (f).} \label{fig:qualit}
\end{figure}

For the sake of increased computational efficiency during simulation, it then
suffices to subtract the baseline velocity off all birds' velocity vectors and
handle displacement rates only. To an observer looking down on the birds
orthogonally to the plane of movement and moving synchronously with them (i.e.,
at the baseline velocity), a sequence of snapshots of the group will seem like
resulting from the birds' movements relative to one another only, even though
they are indeed traveling through space and capable of changing their velocity
vectors, both in magnitude and in direction, through time (forthcoming
Figures~\ref{fig:snap} and~\ref{fig:qualit} show some of these snapshots).

Each simulation runs through $T$ time steps of fixed duration, so the handling
of displacement rates may be further simplified so that only the actual
displacements per time step need in fact be handled. We implement displacements
by simply adding (or subtracting) a fixed amount $\Delta x>0$ (laterally) or
$\Delta y>0$ (longitudinally) to a bird's position per time step.

At each time step, a bird follows Rule~1, if applicable, by altering its
position via $\pm\Delta x$ or $+\Delta y$. If Rule~1 is not applicable, then
applying Rule~2 (and controlling its application via Rule~3) requires a precise
criterion for seeking ``the nearest position that affords an unobstructed
longitudinal view.'' Let us call a gap, in the current time step, any lateral
span that contains the lateral coordinate of no portion of any bird. A gap is
maximal if it is contained in no other gap; maximal gaps, therefore, may be
regarded as inducing longitudinal stripes of empty space, each stripe delimited
by some bird's wing tip on at least one side. Our algorithm implements the
criterion required by Rule~2 as follows. Let $i$ be the bird in question and
consider all maximal gaps that are at least $w+2\lambda$ wide (this includes the
two outermost gaps, which are infinitely wide). The position that $i$ seeks is
given by the nearest wing tip delimiting such a gap, provided the
width-$(w+2\lambda)$ extension of that wing into the gap can be seen by $i$
without obstruction. Aiming at that position, $\pm\Delta x$ is applied as
needed, and eventually Rule~1 may once again become applicable. As for what
remains unspecified of Rule~3, our algorithm implements the lateral separation
by $\lambda$ directly (i.e., Rule~2 is applied if upwash has not been found or
is not optimal).

Collisions are avoided at all time steps whenever attempting a displacement. Two
birds are allowed to get as close to each other as having no lateral separation
between them or a longitudinal separation of $\epsilon$. If the desired
displacement infringes this, then it is not effected and the bird simply chooses
randomly between a longitudinal displacement of $+\Delta y$ or $-\Delta y$,
which is only applied if possible. In our algorithm, this possibility of
displacing a bird longitudinally by $-\Delta y$ represents the only source of
longitudinal deceleration.

\begin{figure}[t]
\centering
\scalebox{0.750}{\includegraphics{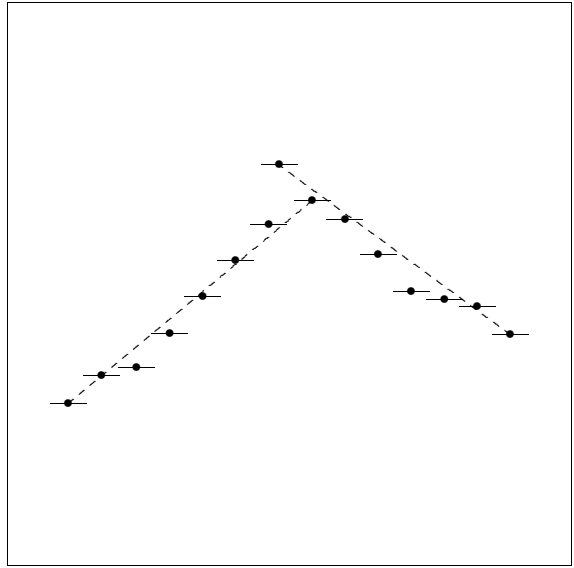}}
\caption{Flight formation with two straight-line segments (shown in dashes).}
\label{fig:lines}
\end{figure}

We have conducted extensive simulations for $n=15,25,35$ and
$\alpha=170^\circ,180^\circ$ (the occurrence of V-like formations in large
flocks is inherently difficult and is rarely observed in nature
\cite{seiler03},\footnote{Except for relatively small birds \cite{anderson04},
in which case V-like formations, as we have noted, seem not to originate
primarily from aerodynamics-related gains. In fact, large formations of
relatively small birds are characterized by lateral separations that are
inherently incompatible with $\lambda$, the optimum predicted by
aerodynamics-based analyses.} thence our moderate choices for the value of $n$).
All simulations used $T=2\,000$ and the values listed in Table~\ref{tab:prms}
for the distance-related parameters. For a quantitative evaluation, we
concentrate on the five indicators listed next, for which results were obtained
as averages over $1\,000$ independent simulations, each one starting at a random
placement of the $n$ birds inside the unit square.
\begin{description}
\item[Time for stabilization:] Number of time steps until all birds stop moving
relative to one another (taken as $T$ for the simulations in which this does not
happen).
\item[Number of lead birds:] How many birds intersect no other's upwash regions.
\item[Number of unconnected groups:] A set of birds is counted as an
unconnected group if no bird outside it intersects any of its own birds' upwash
regions, and conversely [e.g., the situation in Figure~\ref{fig:qualit}(c)
contains two groups].
\item[Number of straight-line segments:] A straight-line segment joins a
trailing bird---one whose upwash regions no other bird intersects---to either a
lead bird or a ``bifurcation'' bird---one whose upwash regions are intersected
by two other birds (e.g., the situation in Figure~\ref{fig:lines} contains two
straight-line segments).
\item[Mean distance to nearest straight-line segments:] Each of the $n$ birds
contributes one distance to the mean, unless it is a lead or ``bifurcation''
bird, in which case it contributes the average distance to the two nearest
segments.
\end{description}
Except for the time for stabilization, all indicators refer to the end of the
simulation.

\begin{figure}[p]
\centering
\begin{tabular}{c@{\hspace{0.20in}}c}
\scalebox{0.315}{\includegraphics{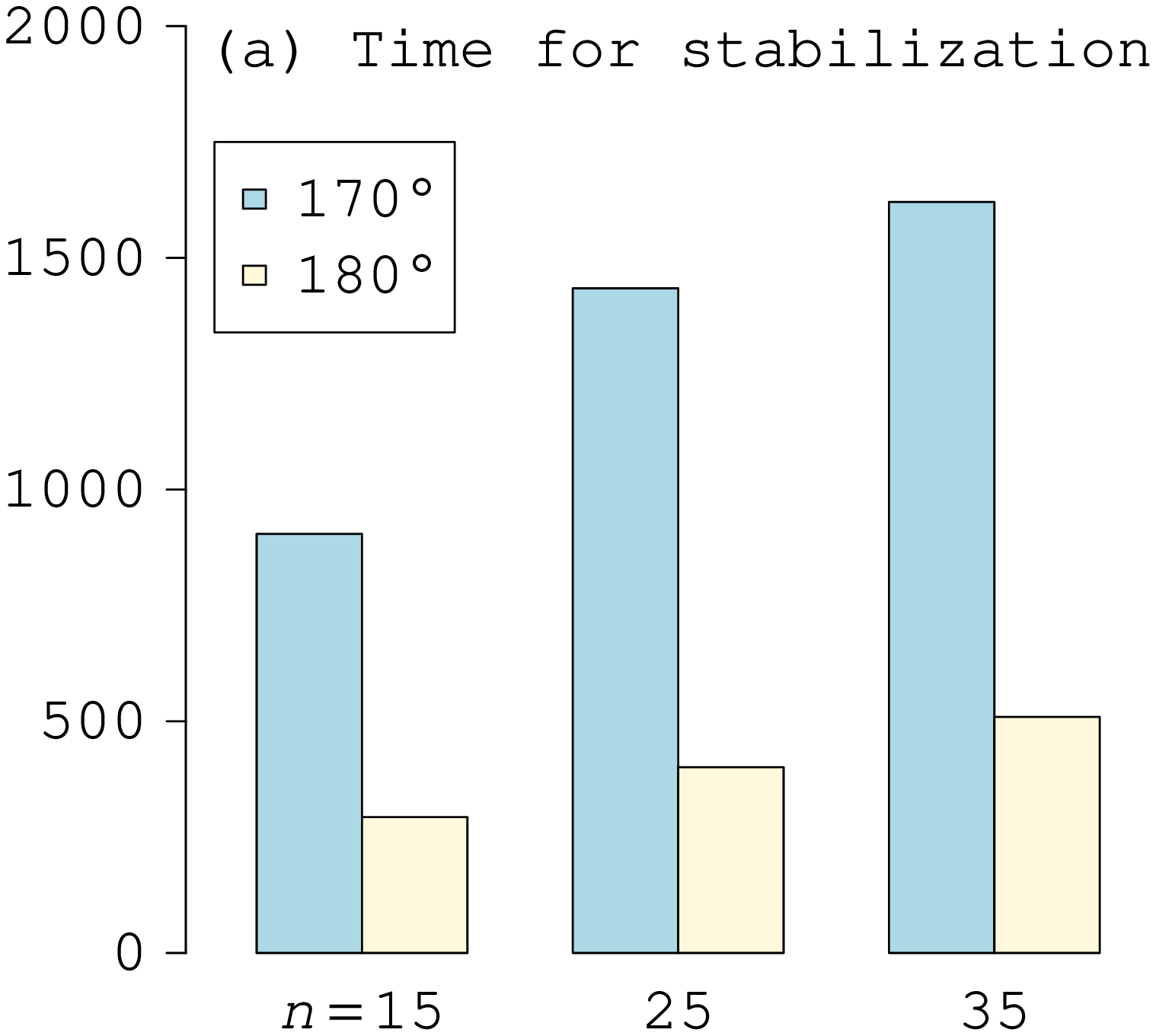}}&
\scalebox{0.315}{\includegraphics{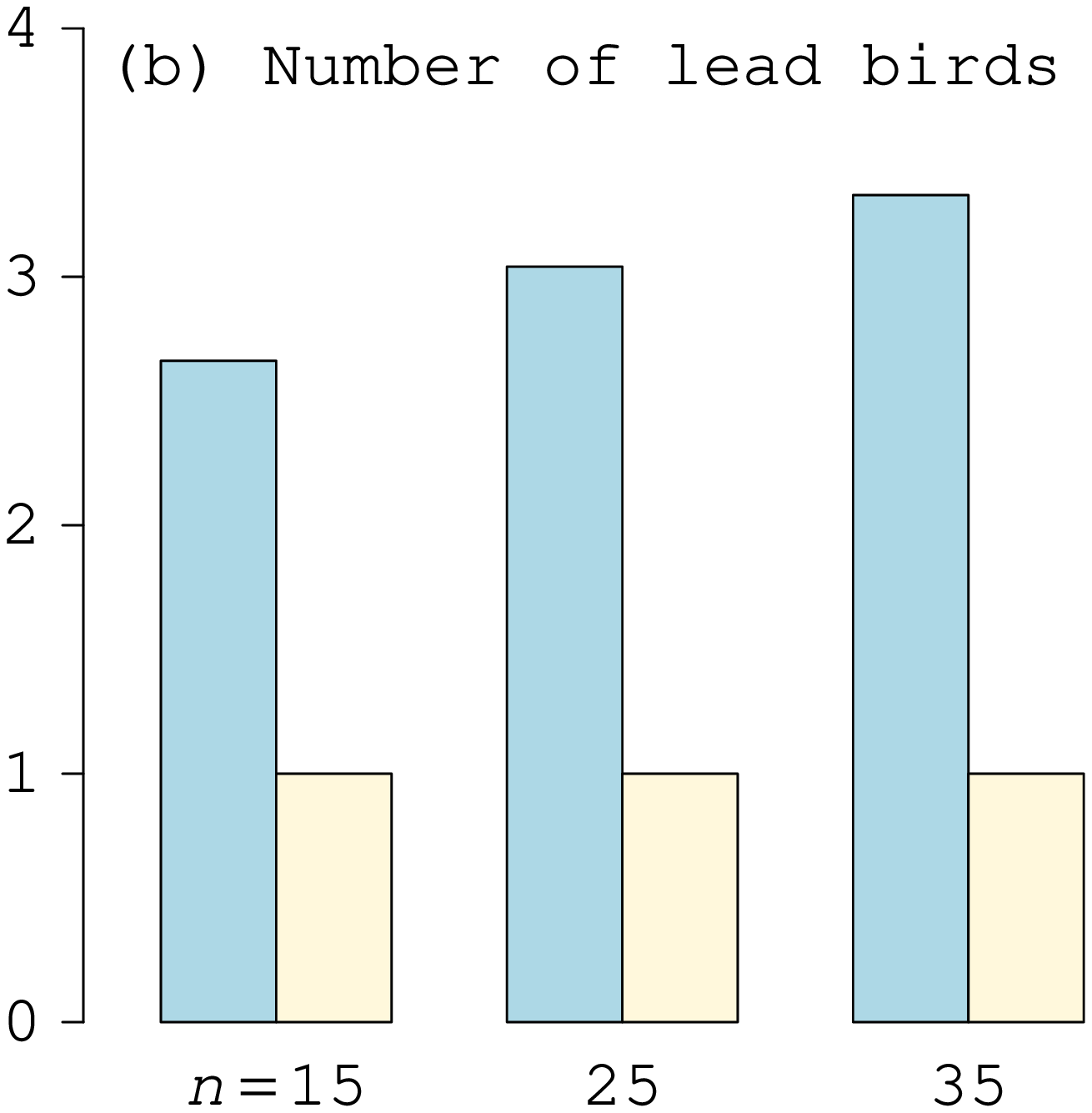}}\\
\scalebox{0.315}{\includegraphics{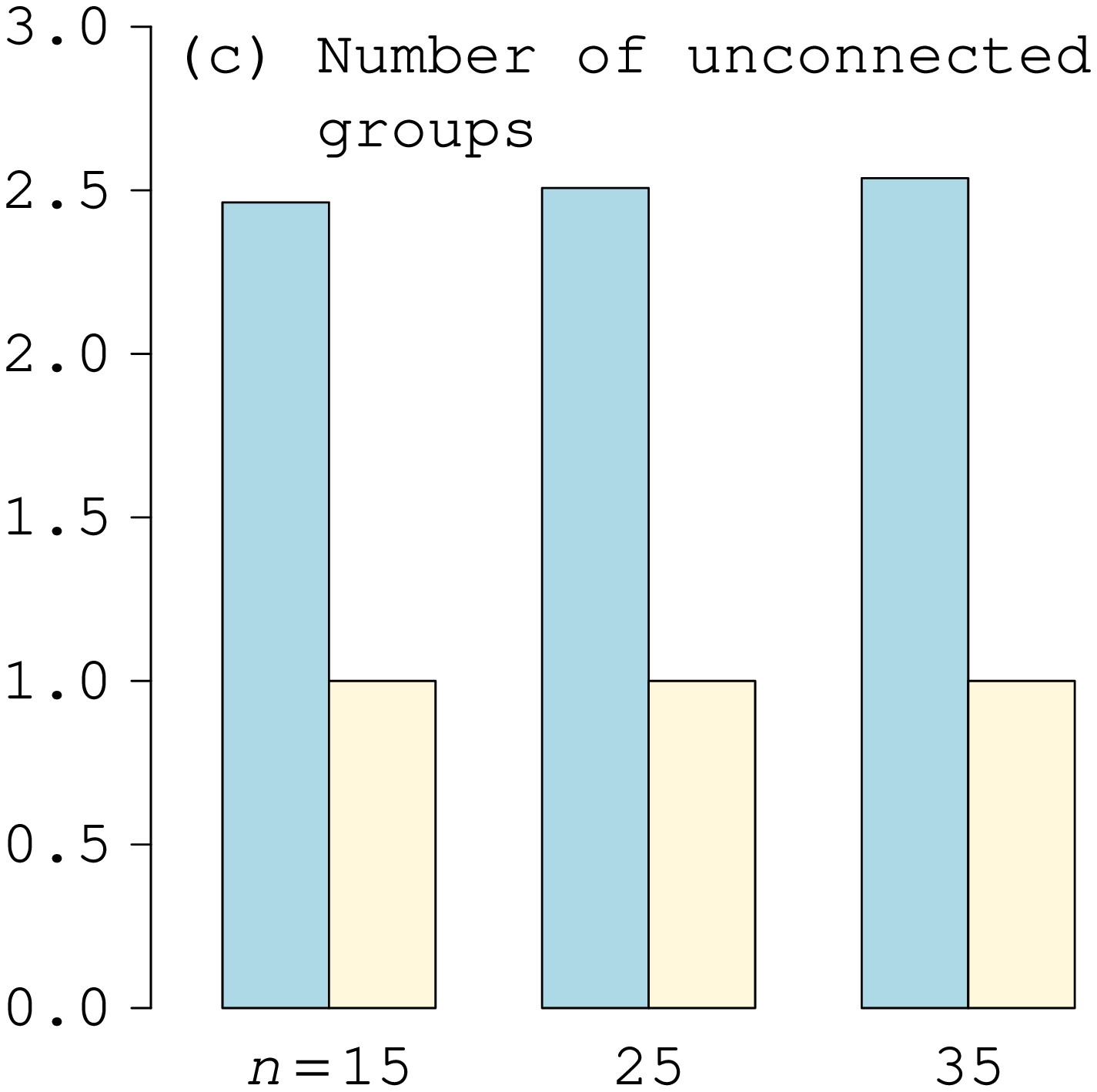}}&
\scalebox{0.315}{\includegraphics{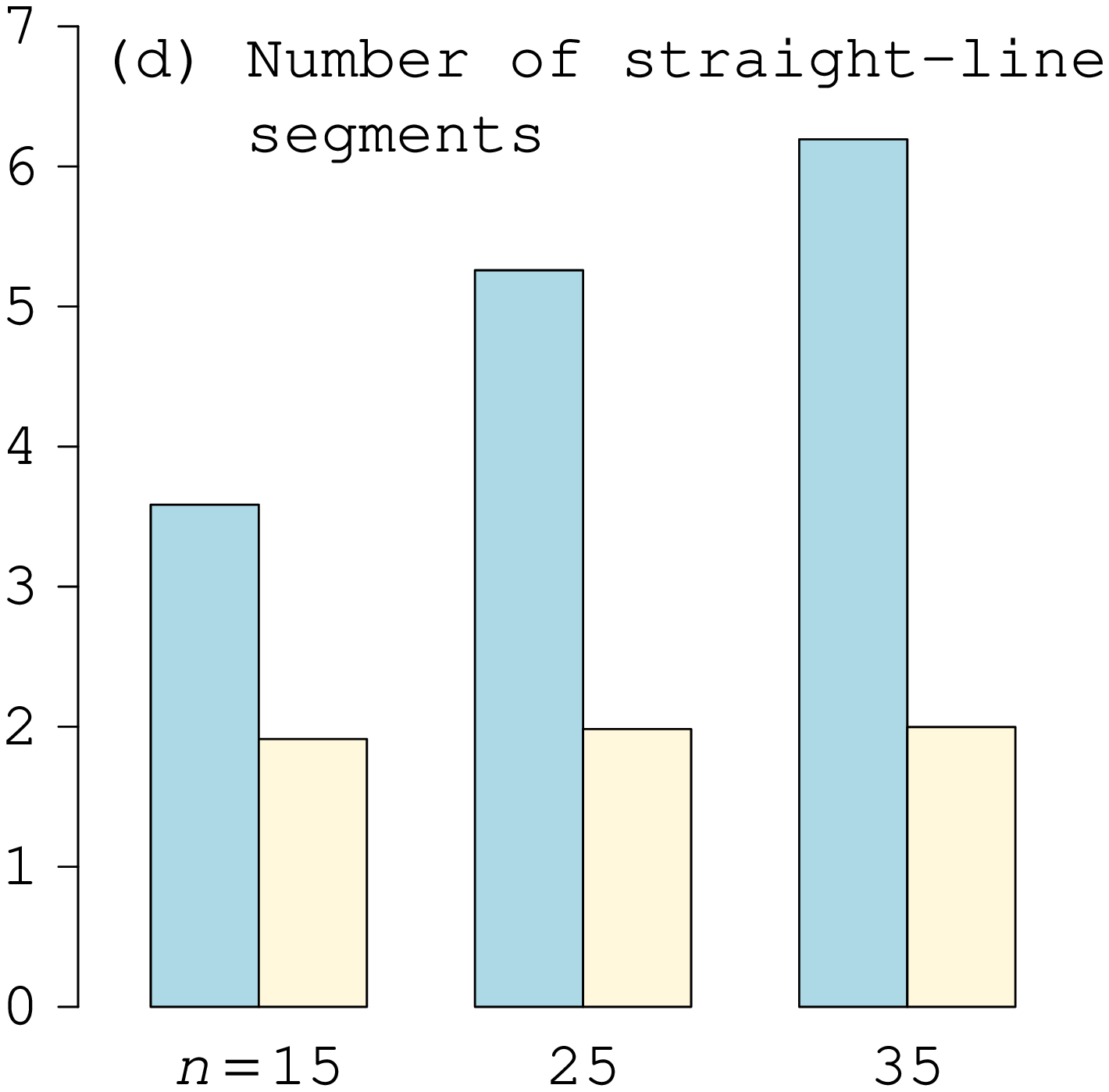}}\\
\scalebox{0.315}{\includegraphics{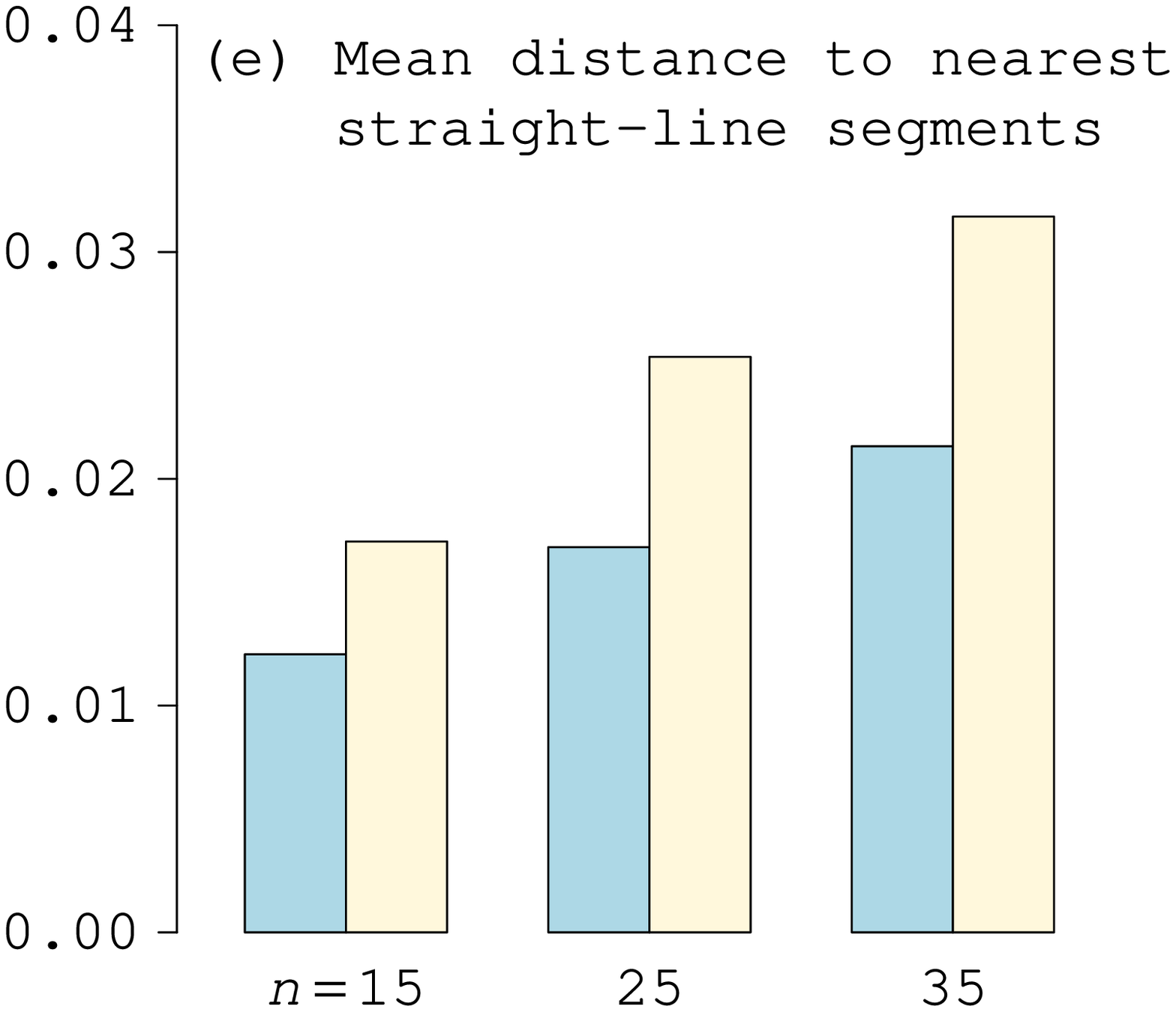}}&
\scalebox{0.315}{\includegraphics{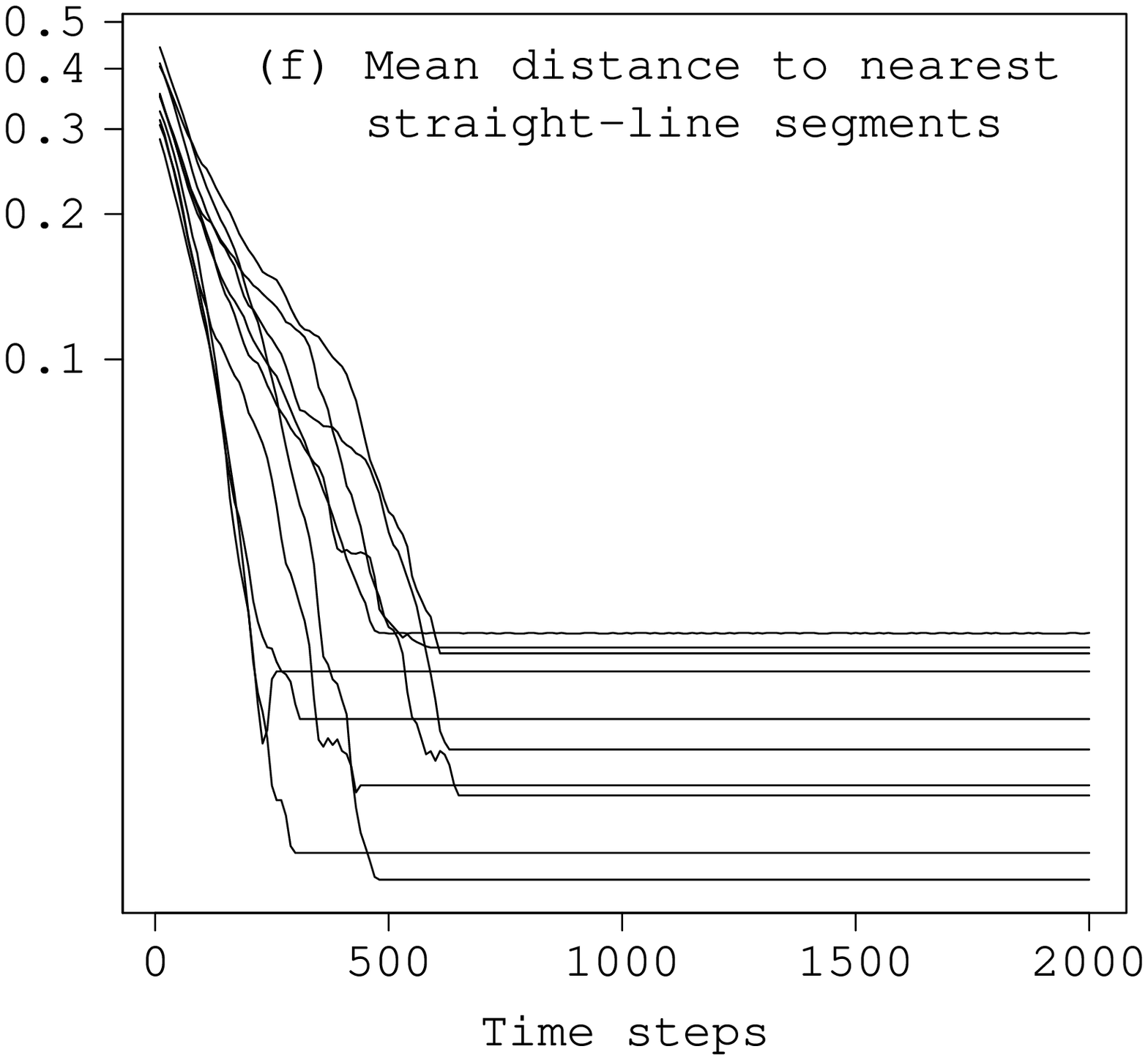}}\\
\end{tabular}
\caption{The five indicators, averaged over $1\,000$ independent simulations
(a--e), and the time evolution of the fifth indicator for ten randomly chosen
simulations with $n=25$ and $\alpha=170^\circ$ (f).}
\label{fig:quantit}
\end{figure}

Two snapshot sequences, in the sense discussed earlier, are shown in
Figures~\ref{fig:snap} and~\ref{fig:qualit}. The former sequence depicts, for
six time steps during the same simulation, the evolution of the birds' positions
from their initial randomly determined situation through the eventual V-like
formation. The latter sequence is a sampler of the formations achieved after $T$
time steps. The averages for the five indicators are shown in
Figure~\ref{fig:quantit}.

\begin{table}[t]
\caption{Parameter values ($\times 1/768$).}
\centering
\begin{tabular}{ccl}
\hline
Parameter&Value&Description\\
\hline
$\ell$&$30$&Lateral size of upwash region (cf.\ Figure~\ref{fig:wash})\\
$d$&$50$&Longitudinal size of upwash region (cf.\ Figure~\ref{fig:wash})\\
$w$&$50$&Wingspan\\
$\Delta x$&$3$&Lateral displacement per time step\\
$\Delta y$&$3$&Longitudinal displacement per time step\\
$\epsilon$&$9$&Margin for longitudinal collision\\
\hline
\end{tabular}

\label{tab:prms}
\end{table}

\section{Discussion and further work}

The plots in Figure~\ref{fig:quantit} reveal that, for $\alpha=180^\circ$, the
average formation into which the $n$ birds settle seems to become stable
relatively quickly, and also to be evocative of the canonical V formation: one
lead bird (hence one single group), and two straight-line segments. Decreasing
$\alpha$ to $170^\circ$ delays stabilization significantly while allowing for an
increased number of lead birds and of unconnected groups, and consequently of
straight-line segments as well.\footnote{We remark that our results for
$\alpha=170^\circ$, when compared to those for $\alpha=180^\circ$, are
qualitatively representative of what we have observed down through roughly
$\alpha=120^\circ$, which singles out $\alpha=180^\circ$ as a sort of special
case.} Behind all these effects are the facts that, with the smaller angle, a
bird tends to have a more limited set of choices for the nearest bird needed in
Rule~1, and also fewer maximal gaps for use in the implementation of Rule~2.
Upon reaching stability, however, the number of birds clustered around each
straight-line segment tends to be smaller, and thus so does the mean distance
between the birds and the nearest segments.

Figure~\ref{fig:quantit}(f) offers the possibility of a closer glimpse into ten
(out of the $1\,000$) simulations for $n=25$ and $\alpha=170^\circ$. The figure
indicates that, even though at this reduced angle stability is on average harder
to achieve, there are simulations in which the mean distance to the nearest
straight-line segments drops nearly vertiginously. This suggests, as in fact we
have confirmed by examining all the data more closely, that the average time for
stabilization is being influenced by some simulations that did not fully
stabilize and contribute each $T=2\,000$ time steps to the average.

While we believe our results to support the view that Rules~1--3 are successful
in helping explain the emergence of V-like formations, we also recognize that
the possibilities for further investigation are very numerous. These include a
more detailed study of how different parameter values in our simulation
algorithm affect the results,\footnote{For example, allowing $\lambda$ to
deviate from its purported optimal value near $-0.1073w$, or yet varying the
value of $d$, is expected to influence the various angles in the resulting
V-like formations.} and also the possibility of studying other algorithmic
approaches to realizing the same set of rules. Some of these other approaches
might, for example, introduce a new parameter to quantify the ``unobstructed
longitudinal view'' mentioned in Rule~2; such a parameter would, essentially,
substitute bounded regions of space for the maximal gaps of our current
algorithm, which are longitudinally unbounded. They might also adopt a
perturbation model to continually effect small fluctuations in the birds'
positions so that stability would never really be achieved; interesting
questions in this case would be whether V-like formations would still occur,
how persistent they would be, and how they would change with time.

To finalize, we remark that Rules~1--3 themselves may be put through ever more
stringent tests. One clear possibility is to consider them in three-dimensional
space and to investigate whether roughly planar V-like formations still emerge.
We find it at this point unclear whether the three rules will hold unchanged,
since a three-dimensional proximity model will be needed for use with Rule~1,
and likewise three-dimensional models of vision obstruction and of flight
aerodynamics for use with the other two rules.

\subsection*{Acknowledgments}

We acknowledge partial support from CNPq, CAPES, and a FAPERJ BBP grant.

\bibliography{biology,engineering}
\bibliographystyle{plain}

\end{document}